\documentclass{sig-alternate-2013}
\usepackage{amsmath}
\usepackage{caption}
\usepackage{subcaption}
\usepackage{multirow}
\usepackage{algorithm}
\usepackage[noend]{algpseudocode}
\usepackage{hyperref}
\usepackage{graphicx}
\usepackage{footnote}
\usepackage{enumitem}
\usepackage{tikz}
\usepackage{booktabs} 
\makesavenoteenv{tabular}
\makesavenoteenv{table}
\makesavenoteenv{table*}

\newcommand*\samethanks[1][\value{footnote}]{\footnotemark[#1]}
\def\x{\mathbf x}
\def\w{\mathbf w}

\begin{document}
\title{Field-aware Factorization Machines \\ in a Real-world Online Advertising System}
 
\numberofauthors{3}
\author{
\alignauthor
Yuchin Juan\thanks{Contributed equally to this work.}\\
\affaddr{Criteo Research}\\
\affaddr{Palo Alto, CA}\\
\email{yc.juan@criteo.com}
\alignauthor
Damien Lefortier\samethanks[1]\\
\affaddr{Facebook}\\
\affaddr{London, UK}\\
\email{dlefortier@fb.com}
\alignauthor
Olivier Chapelle\\
\affaddr{Google}\\
\affaddr{Mountain View, CA}\\
\email{chapelle@google.com}
}

\permission{\copyright 2017 International World Wide Web Conference Committee \\ (IW3C2), published under Creative Commons CC BY 4.0 License.}
\conferenceinfo{WWW'17 Companion,}{April 3--7, 2017, Perth, Australia.}
\copyrightetc{ACM \the\acmcopyr}
\crdata{978-1-4503-4914-7/17/04. \\
http://dx.doi.org/10.1145/3041021.3054185 \\
\includegraphics{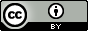}}

\clubpenalty=10000
\widowpenalty=10000
 
\maketitle

\begin{abstract}
Predicting user response is one of the core machine learning tasks in computational advertising.
Field-aware Factorization Machines (FFM) have recently been established as a state-of-the-art method
for that problem and in particular won two Kaggle challenges. This paper presents some results from 
implementing this method in a production system that predicts click-through and conversion rates for display
advertising and shows that this method it is not only effective to win challenges but is also valuable in a 
real-world prediction system. We also discuss some specific challenges and solutions to reduce the
training time, namely the use of an innovative seeding algorithm and a distributed learning mechanism.
\end{abstract}

\section{Introduction}
\label{sec:intro}
Online advertising is a major business for Internet companies and one of the core problem in that field is to be able to match the right advertisement to the right user at the right time.
Accurate click-through rate prediction is essential for solving that problem and has been the topic of extensive research, both 
for search advertising \cite{graepel2010web,mcmahan2013ad} and display advertising \cite{chapelle2014simple,YC16a}.
Performance based advertisers measure the performance of their campaigns not only with respect to clicks, but also to conversions -- defined as a user action on the website such a purchase -- and specific machine learning models have been developed for conversion prediction \cite{lee2012estimating,rosales2012post,chapelle2014modeling, vasile2016cost}.

A prominent model for these prediction problems is logistic regression with cross-features \cite{mcmahan2013ad,chapelle2014simple}. When all cross-features are added, the resulting model is equivalent to a polynomial kernel of degree 2 \cite{chang2010training}. A Kaggle challenge was hosted by Criteo in 2014 to compare CTR prediction algorithms.\footnote{https://www.kaggle.com/c/criteo-display-ad-challenge} Logistic regression with cross-features was indeed 
quite successful in that competition: the 3rd place winner solution was based on this technique \cite{Song:2014}. But the winning solution is  a variant of factorization machines \cite{rendle2012factorization} called {\em Field-aware Factorization Machines} (FFM) \cite{YC16a}. The impressive performance of FFM prompted us to implement it and test it as part of our production system. 
	
\paragraph{FFM}
Consider the case of categorical features -- most features in ad systems are either categorical or can be made categorical through discretization.
Let $F$ be the number of features (or fields) and $v_1,\dots,v_F$ be the values of these features for a given example. The FFM prediction on this example can be written as:\footnote{The prediction here is specific to categorical features while \cite{YC16a} handles the more general case of continuous features.} 
\begin{multline}
\sum_{f_1=1}^F \sum_{f_2=f_1+1}^F \w_{i_1} \cdot \w_{i_2}, \\
\text{where}~~i_1 = \Phi(v_{f_1}, f_1, f_2),~i_2=\Phi(v_{f_2}, f_2, f_1),
\label{eq:ffm}
\end{multline}
with $\w \in \mathbb{R}^{d\times k}$ the weight matrix and $\w_i\in \mathbb{R}^k$ denotes the embedding of the $i$-th entry.
The mapping $\Phi(v, f_1, f_2)$ maps a value $v$ of feature $f_1$ in the context of feature $f_2$ to an index from 1 to $d$.
This may be any hash function or based on dictionary. In the latter case, $d$ will be equal to $F\times \sum_{f=1}^F c_f$, with $c_f$ the cardinality of the $f$-th feature.
	
In regular factorization machines, there is a unique embedding for a given feature value; in other words, the indices in \eqref{eq:ffm} for FM are $i_1=\Phi(v_{f_1}, f_1)$ and $i_2=\Phi(v_{f_2}, f_2)$. But in {\em field-aware} FM, there is a different embedding depending on the other feature of the dot product. As argued in \cite{YC16a} this gives additional modeling flexibility.  

\paragraph{Related work}	 
%As mentioned above, this paper is a follow-up of \cite{YC16a} which itself is a generalization of Pairwise Interaction Tensor Factorization \cite{rendle2010pairwise}. 
A similar effort to ours has been reported by AdRoll in a blog post:\footnote{\url{http://tech.adroll.com/blog/data-science/2015/08/25/factorization-machines.html}} the author reports substantial gains after deploying FMs in their CTR prediction system. 
Google \cite{mcmahan2013ad} and Facebook \cite{he2014practical} may not use FMs but have reported some specific challenges they encountered in productionizing their large scale CTR prediction system, which are related to the challenges for productionizing FFM.
Factorisation Machine supported Neural Network (FNN) and Sampling-based Neural Network (SNN) \cite{zhang2016deep} are two learning algorithms related to FMs that have also been applied a CTR prediction task. They are both deep neural networks but differ in their embedding layer: SNN uses a regular embedding layer while FNN is initialized with the result of a factorization machine.
The recent interest in factorization machines have led to the development of distributed solvers \cite{li2016difacto} for these techniques. Finally a hierarchical version of factorization machines has been introduced in \cite{oentaryo2014predicting}.

Even though FFM have been shown to be a state-of-the-art method for computational advertising by winning two Kaggle challenges, it is still unclear if they are well suited in a production environment. The Netflix challenge is a reminder that a production system has some specific set of constraints and goals that differ from the ones of an academic competition: ultimately Netflix decided not use the winning solution.\footnote{http://techblog.netflix.com/2012/04/netflix-recommendations-beyond-5-stars.html}

This paper discusses our attempt at implementing FFM in a production system that predicts click-through and conversion rates on display advertisements. Section \ref{sec:results} presents offline and online (A/B test) results and provides some insights on the benefits of this method over standard logistic regression as well as the challenges for using FFM in a production system. These positive results further led us to address one of the main bottleneck encountered with our FFM implementation: training speed. 
Section \ref{sec:distributed} investigates how to train FFM in a distributed environment. And Section \ref{sec:seeding} offers an innovative model seeding procedure to further solve that problem, resulting in a more accurate model with a shorter training time and using less computation resources.
Finally Section \ref{sec:conclusion} presents conclusions and future work.

\section{FFM in A Production System}
\label{sec:results}

	In this Section, we describe how we use FFM in our production system, present our offline and online results, and discuss the benefits and challenges of using FFM in such a setting.
	
	\subsection{Baseline}

	As discussed in Section~\ref{sec:intro}, state-of-the-art advertising systems are based on click-through rate (CTR) and conversion rate (CR) prediction models. In this paper, we consider both CTR and CR prediction models used for bidding in real-time auctions (see, e.g., \cite{chapelle2014simple, vasile2016cost}).  To predict the probability of a sale given a display, we use a multiplicative model between a model of the probability of a click given a display and a model of the probability of a sale given a click, as discussed in \cite{chapelle2014modeling}. So, in the rest of the paper we call these two models CTR and CR.
	
	Our baseline system for training these models is based on previous work \cite{agarwal2014reliable, chapelle2014simple, vasile2016cost}. Indeed, following \cite{agarwal2014reliable, chapelle2014simple}, we use the hashing trick \cite{weinberger2009feature} to reduce the dimensionality of our data and to thus reduce the number of parameters to fit. We use logistic regression (LR) with cross-features fitted with L-BFGS warm-started using SGD \cite{agarwal2014reliable, chapelle2014simple}. Following \cite{vasile2016cost}, we also use cost-sensitive learning for the CR model and weight each sale depending on the value of the sale for the advertiser, as this was shown to increase the performance for the CR model both offline and online. We use Hadoop AllReduce for distributing the learning of our models \cite{agarwal2014reliable}.
	
	%\todo{Compare to other systems \cite{mcmahan2013ad,he2014practical,li2015click}?} Compared to DiFacto \cite{li2016difacto}, we use the hashing trick to reduce the dimensionality of the data instead of memory adaptive constraints and a sparse regularization--- as they do. \todo{Explain difference/why.}
	
	Below, we investigate the usage of FFM instead of LR for training our CTR and CR prediction models. We still use the hashing trick. So, the mapping $\Phi(v, f_1, f_2)$ in \eqref{eq:ffm} is based on a hash which a fixed hashing space (of the order of tens of millions).

    \subsection{Offline comparison}

    We now present results comparing FFM to the state-of-the-art baseline on an offline dataset.
	
		\paragraph{Offline metrics} 
		
		We use two offline metrics. First, we use the normalized log loss (NLL). This metric shows the relative improvement in log loss (LL) of the model to be evaluated versus a baseline predictor, in our case the average empirical CTR or CR of the dataset, similar to the normalization in \cite{he2014practical,lefortier2015learningsys,vasile2016cost}. This metric is defined formally for any prediction $p$ as follows, where we denote $\bar{p}$ the best constant predictor on the test set and $N$ the number of impressions in our dataset.

\begin{equation}
\textbf{LL}(p) = -\sum_{i=1}^{N} y_i \log(p_i) + (1 - y_i) \log(1-p_i)  \\
\label{eq:msew}
\end{equation}

\begin{equation}
\textbf{NLL}(p) = \frac{\text{LL}(\bar{p}) - \text{LL}(p)}{\text{LL}(\bar{p})} \\
\label{eq:nll}
\end{equation}

We also use the \emph{Utility}\footnote{This metric is called {\em expected} Utility in \cite{chapelle2015offline}, but we refer to it as Utility in this paper.} metric \cite{chapelle2015offline,vasile2016cost}, which allows to model offline the potential change in profit due to a prediction model change. Since the observed profit in historical data is fixed, this metric makes the assumption that the display costs are determined by the highest second bids coming from a second price auction and that they are generated according to a distribution conditioned on the observed display cost. This metric is defined as follows, where $v_i$ is the reward of the $i^{th}$ impression.
\begin{equation}
\textbf{Utility} = \sum_{i}\int_{0}^{p(\x_i)v_i} (y_i \cdot v_i - \tilde{c})\Pr(\tilde{c} \mid c_i) d\tilde{c} \\
\label{eq:eu}
\end{equation}
%\text{EU} = \int_0^{pv} (yv-\tilde{c}) \Pr(\tilde{c} \mid c) d\tilde{c}

The distribution $\Pr(\tilde{c} \mid c)$ specifies what could have been the second price instead of the observed cost $c$; \cite{chapelle2015offline} suggests a Gamma distribution with $\alpha=\beta c + 1$  and free parameter $\beta$. The motivation for selecting this distribution is that it interpolates nicely between two limit distributions: a Dirac distribution centered at $c$ (as $\beta \to +\infty$) and a uniform distribution (as $\beta \to 0$). It can be shown that the utility with a uniform distribution is equivalent to a weighted squared error \cite{hummel2013loss}.
	
	\paragraph{Experimental setup}
	
	We use internal data from Criteo to do our experiments. Note, however, that, as discussed in Section~\ref{sec:intro}, FFM have been shown to be better than existing methods on many public data sets already \cite{YC16a}. Moreover, the goal of this section is to show we can improve upon our baseline using FFM in a real-world online advertising system, which uses its own data. We need offline experiments to ensure that FFM are performing well in our system (both in terms of predictive performance and scalability) and for parameter tuning, before we can perform a live experiment (A/B test).
	
	We use a variant of progressive validation, similar to \cite{mcmahan2013ad}, for our experiments. The day following the training period serves as a validation set. As shown on Figure \ref{fig:TrainingProcess} below, the process is repeated $N$ times, shifting the learning period (indicated by "tr") by 1 day at each step. The final results are the average metrics over all the test sets (indicated by "te"). 

    \begin{figure}[h]
	    \centering
        \begin{tikzpicture}
            %\node at (0, 0.5) {0am};
            %\node at (1, 0.5) {3am};
            %\node at (2, 0.5) {6am};
            %\node at (3, 0.5) {$\cdots$};

            \draw [fill=gray] (0, 0) rectangle (4.0, -0.5);
            \node at (2.0, -0.25) {tr \#1};
            \draw             (4.0, 0) rectangle (5, -0.5);
            \node at (4.5, -0.25) {te \#1};

            \draw [fill=gray] (1, -0.5) rectangle (5.0, -1.0);
            \node at (3.0, -0.75) {tr \#2};
            \draw             (5.0, -0.5) rectangle (6, -1.0);
            \node at (5.5, -0.75) {te \#2};

            \node at (3.0, -1.25) {$\vdots$};

            %\draw [fill=gray] (3, -1.0) rectangle (6.5, -1.5);
            %\node at (4.75, -1.25) {train \#3};
            %\draw             (6.5, -1.0) rectangle (8, -1.5);
            %\node at (7.25, -1.25) {test \#3};
        \end{tikzpicture}
        \caption{Progressive validation}
        \label{fig:TrainingProcess}
    \end{figure}
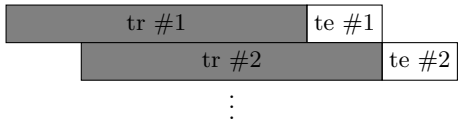

	Parameter tuning is done on a separate temporal slice of data from the data of \ref{fig:TrainingProcess} used for the final experiments. Following \cite{YC16a} the following parameters are tuned: the regularization parameter, the learning rate, and the number of latent factors. We use early-stopping to avoid over-fitting.

	Below and in the rest of the paper, we use confidence intervals computed using bootstraps \cite{efron1994introduction} at the 90\% level. Finally the learning of FFMs is multi-threaded as in \cite{YC16a} to reduce the learning time.

	\paragraph{Latency \& memory consumption}
	
		One potential drawback of using FFM in a production system is that they require more CPU time for inference \cite{YC16a}. This may lead to increased latency online when responding to bid requests and therefore come more timeouts. FFM also require more memory for storing the model as the number of latent factors and/or the number of fields increase, which may lead to a much larger memory consumption than LR.
		
		To solve the memory issue, we propose to reduce the size of the hashing space of FFM models (compared to our baseline) so that FFM models have the same size as the LR models (the exact value depends on the number of fields and on the number of latent factors). Note that if we had not reduced the size of the hashing space, but kept it constant, the size of FFM models would be more than a \emph{100 times} larger than our baseline, which would make it impractical. Therefore, in the results below, FFM and LR models have the same number of parameters (unlike in \cite{YC16a}).

To solve the latency issue, we propose to reduce the number of latent factors as much as possible without significantly degrading the performance of FFM.
Using these two solutions, FFM and LR consume the same amount of memory and we can limit the impact on latency to handle the requirement of our production system, as we will see below.

		\begin{table*}[t]
\centering
\caption{Offline relative comparison between Logistic Regression (baseline) and FFM on our CTR and CR prediction tasks in terms of the NLL metric \eqref{eq:nll}. We present results on all advertisers and on small advertisers -- defined as advertisers with less than 30 sales per day on average. NLL of the CTR (resp. CR) model is the NLL of the probability of a click (resp. sale) given a display.  Statistical significance is indicated by $\blacktriangle$.}
\label{table:offline_results_nll}
\begin{tabular}{lccrc}
   \toprule

\multirow{2}{*}{ Prediction model with FFM} & NLL &  NLL  \\
    & on all advertisers  &  on small advertisers  \\
   \midrule
   CTR & \textbf{$+3.71\%$}$\blacktriangle$ & \textbf{$+5.9\%$}$\blacktriangle$   \\
   CTR + CR & \textbf{$+1.21\%$}$\blacktriangle$ & \textbf{$+6.2\%$}$\blacktriangle$  \\
   \bottomrule
   \end{tabular}
\end{table*}

		\begin{table*}[t]
\centering
\caption{Offline relative comparison between Logistic Regression (baseline) and FFM on our CTR and CR prediction tasks in terms of Utility metrics \eqref{eq:eu}. We report the Utility of our model for the expected number of sales given display, which uses our CTR and CR models as sub-models.  Statistical significance is indicated by $\blacktriangle$.}\label{table:offline_results_utility}
\begin{tabular}{lcccc}
   \toprule
\multirow{2}{*}{ Prediction model with FFM} & Utility$_{\beta=10}$ & Utility$_{\beta=10}$ &Utility$_{\beta=1000}$ & Utility$_{\beta=1000}$  \\
    & on all advertisers  &  on small advertisers  & on all  advertisers  &  on small advertisers  \\

   \midrule
   CTR & \textbf{$+6.29\%$}$\blacktriangle$ & \textbf{$+9.70\%$}$\blacktriangle$ & \textbf{$+2.22\%$}$\blacktriangle$  & \textbf{$+4.39\%$}$\blacktriangle$  \\
   CTR + CR & \textbf{$+11.42\%$}$\blacktriangle$ & \textbf{$+38.44\%$}$\blacktriangle$ & \textbf{$+5.43\%$}$\blacktriangle$  & \textbf{$+18.34\%$}$\blacktriangle$ \\
   \bottomrule
   \end{tabular}
\end{table*}
		
	\paragraph{Offline results}

We compare LR and FFM on our CTR and CR prediction tasks in terms of NLL (Table \ref{table:offline_results_nll}) and Utility (Table \ref{table:offline_results_utility}). 
FFM achieves significantly better results with a large effect compared to LR, both in terms of NLL and of Utility for our CTR model, thus confirming the results from \cite{YC16a} on our data. We also observe large gains on our CR model, thus extending the results from \cite{YC16a} to CR models on all our offline metrics. 

We also observe that the improvements are even larger on small advertisers, which represent a significant portion of our traffic, for both our CTR and CR models on all metrics.
Our hypothesis to explain these results has to do with sparse data and unobserved cross-features: LR is unable to predict the value associated with a cross-feature that is not part of the training data; on the other hand, FFMs are able to better generalize through their latent representation (see detailed explanation and example in \cite[Section 2]{YC16a}). For large advertisers LR has enough data to learn a good model, but for small advertisers FFMs handle this data sparsity issue better than LR.
	
	During the tuning of the hyper-parameters, we observed very similar results as in \cite{YC16a} in terms of performance w.r.t each hyper-parameter. The most important parameter is the number of epochs and we use early-stopping to automatically tune it. 
	
	We also investigated the prediction time of FFM compared to the baseline model, which is expected to increase despite the fact that we constrained our FFM models to be of the same size of the baseline. This is because the number of operations to compute the prediction \eqref{eq:ffm} is $O(F^2k)$ while LR with all cross-features requires only $O(F^2)$ operations. We observed that the slowdown of FFM is indeed proportional to the number of latent factors $k$. It turns out that $k=2$ is a good trade-off: it hardly degrades the accuracy results compared to the results above which were obtained with 4 latent factors (0.1\% in NLL) and a 2x in prediction time in our system  is acceptable since prediction is is not the most time-consuming part of processing a request (compared to extracting raw features, pre-processing them, etc.).

    \subsection{Online comparison}

		\begin{table*}[t]
\centering
\caption{Online relative comparison between Logistic Regression (baseline) and FFM on our CTR and CR prediction models in terms of Return On Investment (ROI), i.e. advertiser value over cost,  during our A/B test. Statistical significance is indicated by $\blacktriangle$.}
\label{table:online_results}
\begin{tabular}{lcc}
   \toprule
\multirow{2}{*}{ Prediction model with FFM} & ROI &  ROI   \\
    & on all advertisers  &  on small advertisers   \\
   \midrule
   CTR + CR & \textbf{$+0.97\%$}$\blacktriangle$ & \textbf{$+2.61\%$}$\blacktriangle$  \\
   \bottomrule
   \end{tabular}
\end{table*}
		
	As the offline results were quite promising, we decided to run an A/B test using FFM for both CTR and CR predictions models. Although FFM require more time for the inference (see above), we did not observe any significant impact on our timeouts while serving live traffic. So, we were able to A/B test FFM on a large portion of our live traffic. This A/B test served $\sim$5B displays ($\sim$2.5B for each population).
	
	During the A/B test, we ensured that both the baseline model and FFM were refreshed online \emph{synchronously} since different refresh rates might bias the results. Even with multi-threading, the learning time of FFM is indeed much higher than for our distributed optimization baseline. In Section~\ref{sec:distributed} and \ref{sec:seeding}, we will see how to reduce this learning time, but now we focus only on the performance improvements we can get online with FFM. The results we obtained are the following. 

	Results are shown in Table~\ref{table:online_results}. We observed an increase in the number of displays ($+4.59\%$), while the overall display cost stayed almost constant. We observed less clicks, but more conversions leading to more advertiser value for the same cost. Our change therefore resulted in a significant positive impact: +0.97\% of  Return On Investment (ROI), that is of advertiser value over cost, which is substantial.
	We also observed that the improvements were even larger on small advertisers (defined as the ones with less than 30 sales per day), which represent a significant portion of our traffic. On small advertisers, we observed an increase in the number of displays ($+4.85\%$), while the overall display cost stayed almost constant too. We also observed less clicks, but even more conversions leading to even more advertiser value for the same cost and to +2.61\% of  Return On Investment (ROI), which is remarkable. 

This confirms our offline results and shows that one of the strengths of FFM is indeed their ability to generalize better than logistic regression through their use of a latent representation.
	
	\subsection{Discussion}
	
	Our positive online results motivate us to use FFM in production instead of LR. To do so, the code change is rather small if SGD is already available. However, there are a few challenges to keep in mind when using FFM instead of LR in a production system.
	
	The main concern with rolling out FFM is the learning time, which is much higher than the baseline as discussed before. This means that our models would be refreshed less often with FFM, at the cost of reducing the performance of the system. All our offline experiments to improve our models would also take much longer. This is \emph{not acceptable} and we will discuss in the next two sections how to tackle this problem to handle the scale of a large production system, in particular by distributing the learning on multiple machines.
	
	There are also other challenges. Above, we discussed the memory consumption and prediction latency issues and we showed how to manage them. Another potential problem is the non-convexity of the objective function of FFM, which may lead to some instability in the performance of FFM due to local minimums. To investigate this, we learned multiple FFMs on the same dataset initialized with random weights as in \cite{YC16a}. We observed that all the models have similar performance ($\pm 0.05\%$ of NLL) despite the different initializations. The local minimum issue is thus not a major concern.

	We also saw above that the number of hyper-parameters in FFM is larger than for LR with the addition of the learning rate (as we use L-BFGS for training our LR models) and of the number of latent factors, while we only had the regularization parameter to tune for LR. This means that tuning takes more time when improving our models. However, and as discussed in \cite{YC16a}, this is not a major problem for multiple reasons. First, the performance is not very sensitive to the number of latent factors and to the regularization parameter, while a good value for the learning rate is easy to find. We also found the performance of FFM to be stable over time w.r.t to the hyper-parameters (no need for constant re-tuning).

	As we  have not been able to find a satisfying regularizer for FFM, we use early-stopping to avoid over-fitting \cite{YC16a}--- the only solution we have. So, some monitoring should also be added to ensure that we are not under-fitting or over-fitting despite using early stopping (e.g., if the small amount of data used for testing and deciding when to stop is not representative). 
		
	Note finally that for efficient regression testing \cite{lefortier2015learningsys}, we need to fix the seed used for randomizing the initial weights \cite{YC16a}.

\section{A Simple Distributed Setting}
\label{sec:distributed}

    In the previous section, we discussed that the training time of FFM is too slow to meet our production requirement, even after applying the parallelization approach mentioned in \cite{YC16a} on a multi-core machine.
	
    To get more speed-up, a natural option is to train FFM on a distributed system.
    Generally speaking, for sequential algorithms such as SGD or dual coordinate descent, the convergence of their parallelization depends on how often each worker can access the model.
    In shared-memory systems, because each thread can access the model in real-time, it is possible that the convergence remains the same, as shown in \cite{YC16a}.
    However, in distributed systems, where we need to use the network for communication, we can no longer share the model among machines in real-time (due to network overhead). There are two main ways of distributing a stochastic gradient algorithm, {\em synchronously} and {\em asynchronously}. In both cases, each machine has a subset of the data and its own local model and it updates the global model after a batch of data points has been processed. 
    The asynchronous training is often referred to as the {\em parameter server} approach \cite{li2014paramserver, li2016difacto, jeffrey2016distblief}: some machines are dedicated to storing the global model and the workers are continuously reading and updating that model with their local model.
    The synchronous training on the other hand is referred to as {\em iterative parameter mixing} (IPM) \cite{mcdonald2010ipm, martin2010psgd, agarwal2014reliable}: all the models are averaged after a certain amount of data has been processed (e.g. every epoch). 
    From an engineering point of view, simplicity is one of the most important factors we consider when choosing an algorithm. 
    A complicated algorithm requires more time for development, is harder to maintain, and is more likely to introduce bugs. 
    Therefore, in practice, if a simpler algorithm can solve our problem, we would not go for a more complicated one. 
    As we will see, with IPM we are already be able to speed-up the training time 12x with 32 machines.
    This already meets our requirement, so we do not investigate the parameter server approach in this paper.
    IPM for the AdaGrad learning algorithm \cite{JD10a-short} is described in Algorithm \ref{alg:psgd}.
    \begin{algorithm}[t]
        \caption{Iterative Parameter Mixing (IPM) for AdaGrad}
        \label{alg:psgd}
        \begin{algorithmic}[1]
            %\Require $Z$: user-specified outer iterations 
            \State Split $m$ data points across $k$ machines
            \State Initialize $\boldsymbol{w}$
            \State Initialize $G_i \gets I \quad \forall i \in \{1, \cdots, k\}$
            \For{$t \in \{1, \cdots, T\}$} \Comment $T$: number of epochs
                \State Let $\boldsymbol{w}_i \gets \boldsymbol{w} \quad \forall i \in \{1, \cdots, k\}$
                \For{$i \in \{1, \cdots, k\}$ {\em parallel}}
                    \For{each data point}
                        \State Calculate the gradient $\boldsymbol{g}$
                        \State Update $G_i$: $G_i \gets G_i + \text{diag}(\boldsymbol{g}\boldsymbol{g}^T)$
                        \State Update $\boldsymbol{w}_i$: $\boldsymbol{w}_i \gets \boldsymbol{w}_i - \eta G_i^{-1/2} \boldsymbol{g}$ 
                    \EndFor
                \EndFor
                \State $\boldsymbol{w} \gets \sum_{i=1}^k \boldsymbol{w}_i / k$
            \EndFor
        \end{algorithmic}
    \end{algorithm}

    \par
    The speed-up of a distributed algorithm can be modeled by the following equation
    \begin{equation*}
        \text{speed-up} = \text{\#machines} \times \frac{\text{\#epochs with multiple machines}}{\text{\#epochs with one machine}}
    \end{equation*}
    This equation is based on two assumptions:
    \begin{enumerate}
        \item Each machine finishes the computation at almost the same time.
        \item The communication cost among machines is negligible.
    \end{enumerate}
    In our case, both assumptions indeed hold. The first assumption holds, because we equally distribute the training data to all machines and make sure that each machine has similar computing power. 
    The second one holds because IPM only requires synchronization at the end of each epoch, making the synchronization time much less than computation time.
    \par
    The ``real'' distributed algorithm is embedded in our internal system and run on our internal datasets, therefore we are not able to release it.
    For experimental reproducibility, in this paper we use multi-threading to simulate machines, and use the dataset obtained from Criteo's CTR Prediction Challenge described in Section \ref{sec:intro}.
    This simulation is close to reality because the speed-up of a distributed algorithm depends only on the number of machines used and on the slow down in convergence, which can be exactly simulated by multi-threading.

    \par
    If we directly apply IPM, then the convergence gets slower and slower when we keep adding machines.
    The experiment result is shown in Table \ref{tab:DistributedS}.
    Suppose we use 32 machines instead of 1, although the computation is 32 times faster, it also needs 20 times more epochs.
    Therefore, the speed-up is only $32 / (157 / 8) \approx 1.6$.
    A natural way to make the convergence faster is to increase the learning rate $\eta$.
    Though increasing the learning rate indeed makes the algorithm converge faster, it also make the log loss worse.
    This result is shown in Table \ref{tab:DistributedR}.
    \begin{table}[t]
        \centering
        \begin{tabular}{l|rr}
        \# machines& \#epochs & log loss \\
        \hline
        1 & 8 & 0.44552 \\
        2 & 15 & 0.44548 \\
        4 & 29 & 0.44549 \\
        8 & 47 & 0.44560 \\
        16 & 100 & 0.44554 \\
        32 & 157 & 0.44585 \\
        \end{tabular}
        \caption{The number of epochs required to reach the best log loss with different number of machines. Algorithm \ref{alg:psgd} is applied. The learning rate $\eta$ is 0.2.}
        \label{tab:DistributedS}
    \end{table}

    \begin{table}[!htb]
        \begin{subtable}{.5\linewidth}
            \begin{tabular}{l|rr}
            $\eta$ & \#epochs & log loss \\
            \hline
            0.2 & 157 & 0.44585 \\
            0.5 & 70 & 0.44569 \\
            1.0 & 37 & 0.44590 \\
            2.0 & 26 & 0.44622 \\
            3.0 & 21 & 0.44654 \\
            4.0 & 19 & 0.44688 \\
            5.0 & 18 & 0.44721 \\
            \end{tabular}
        \caption{Algorithm \ref{alg:psgd}}
        \label{tab:DistributedR}
        \end{subtable}%
        \begin{subtable}{.5\linewidth}
            \begin{tabular}{l|rr}
            $\eta$ & \#epochs & log loss \\
            \hline
            0.2 & 200 & 0.44819 \\
            0.5 & 130 & 0.44600 \\
            1.0 & 55 & 0.44578 \\
            2.0 & 31 & 0.44565 \\
            3.0 & 22 & 0.44577 \\
            4.0 & 18 & 0.44592 \\
            5.0 & 16 & 0.44608 \\
            \end{tabular}
        \caption{Algorithm \ref{alg:psgd++}}
        \label{tab:DistributedA}
        \end{subtable}
        \caption{With 32 machines, the number of epochs required to reach the best log loss with different learning rates.}
    \end{table}

    \par
	
	We propose the following approach to solve this issue. 
    Remember that, following \cite{YC16a}, we use AdaGrad \cite{JD10a-short} to boost the performance of SGD.
    AdaGrad records the squared gradient sum ($G$) to dynamically adjust the learning rate for each dimension.
    In Algorithm \ref{alg:psgd}, $G$ is not synchronized among machines.
    It may make $G$ on each machine very small and make the effective learning rate too large.
    Based on an idea similar to \cite{agarwal2014reliable}, we aggregate $G$ among each machine at the end of each epoch.
    This new algorithm is described in Algorithm \ref{alg:psgd++}.
    The experiment result is shown in Table \ref{tab:DistributedA}.
    The log loss is much better when a large learning rate is used.
	\begin{algorithm}[t]
    \caption{Improved IPM for AdaGrad}
    \label{alg:psgd++}
    \begin{algorithmic}[1]
        %\Require $Z$: user-specified outer iterations 
        \State Spread $m$ data points into $k$ machines
        \State Initialize $\boldsymbol{w}$
        \State Initialize $G \gets I$
        \For{$t \in \{1, \cdots, T\}$} \Comment $T$: number of epochs
            \State Let $\boldsymbol{w}_i \gets \boldsymbol{w} \quad \forall i \in \{1, \cdots, k\}$
            \State {\color{red} Let $G_i \gets G \quad \forall i \in \{1, \cdots, k\}$}
            \For{$i \in \{1, \cdots, k\}$ {\em parallel}}
                \For{each data point}
                    \State Calculate the gradient $\boldsymbol{g}$
                    \State Update $G_i$: $G_i \gets G_i + \text{diag}(\boldsymbol{g}\boldsymbol{g}^T)$
                    \State Update $\boldsymbol{w}_i$: $\boldsymbol{w}_i \gets \boldsymbol{w}_i - \eta G_i^{-1/2} \boldsymbol{g}$ 
                \EndFor
            \EndFor
            \State $\boldsymbol{w} \gets \sum_{i=1}^k \boldsymbol{w}_i / k$
            \State {\color{red} $G \gets \sum_{i=1}^k G_i$}
        \EndFor
    \end{algorithmic}
\end{algorithm}

    \par
    Under this setting, if we choose $\eta=3.0$, the speed-up we can achieve is $32 \times (8 / 22) \approx 12$.
    Indeed, after we applied this setting in our system, we observe a similar speed-up, which enables us to train a model as fast as our current system.

\section{Warm-start}
\label{sec:seeding}

    As described in Section \ref{sec:results}, we regularly re-train models.
    In Figure \ref{fig:TrainingProcess}, suppose each training set contains several days of data, and we move a few hours forward at each step, there will be a large amount of overlap between training sets \#1 and \#2.
    This means that the model obtained from \#1 may be very similar to one obtained from \#2.
    For logistic regression, the training time to obtain model \#2 can be significantly reduced by initializing model \#2 with model \#1. 
    This technique is known as warm-start \cite{BYC15a-short, CHT14a-short, DD00a-short}.
	
	    \begin{algorithm}[t]
        \begin{algorithmic}
        \Require an initial model ${\bf w}_0$
          \State ${\bf w} \gets {\bf w}_0$
          \State calculate the validation loss $L_0$
          \For{$t \in \{1, \dots, T\}$}
            \State update ${\bf w}$
            \State ${\bf w}_t \gets {\bf w}$
            \State calculate the validation loss $L_t$
            \If{$L_t > L_{t-1}$}
              \State \Return ${\bf w}_{t-1}$
            \EndIf
          \EndFor
        \end{algorithmic}
        \caption{A naive warm-start}
        \label{alg:NaiveSeeding}
    \end{algorithm}

    \par
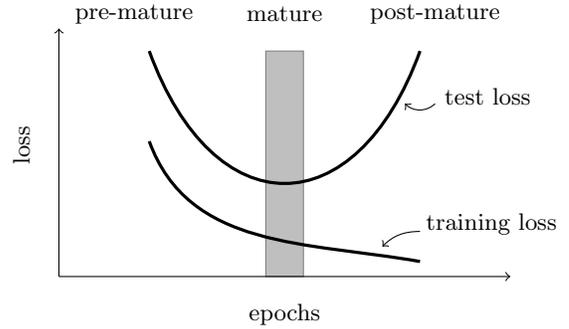
\begin{figure}[t]
        \centering
        \begin{tikzpicture}
            \draw[fill=gray, opacity=0.5] (1.75, 0) rectangle (2.25, 3);
            \draw[->] (-1,0) -- (5, 0);
            \draw[->] (-1,0) -- (-1, 3.3);
            \node at (2, -0.5) {epochs};
            \node[rotate=90] at (-1.5, 1.75) {loss};
            \node at (0, 3.5) {pre-mature};
            \node at (2, 3.5) {mature};
            \node at (4, 3.5) {post-mature};

            \node at (4.7, 2.4) {test loss};
            \draw[->, out=-135, in=-45] (4.0, 2.3) to (3.6, 2.3);

            \node at (4.75, 0.7) {training loss};
            \draw[->, out=180, in=45] (3.8, 0.6) to (3.3, 0.4);
            %\node[rotate=-90] at (-1.5, 1) {validation loss};
            \draw[out=-70, in=-110, distance=2.5cm, very thick] (0.2,3) to (3.8,3);
            \draw[out=-70, in=170, very thick] (0.2,1.8) to (3.8, 0.2);
        \end{tikzpicture}
        \caption{An illustration of over-fitting problem.}
        \label{fig:Overfitting}
    \end{figure}

    For logistic regression, a convex optimization problem, the model will eventually converge to the global optimum no matter warm-start is used or not.\footnote{Assuming an appropriate optimization method and a tight stopping criteria are applied.}
    Warm-start only influences the convergence speed.
    However, this is not the case for FFM.
    To explain why, we first review an undesired property of FFM that has been investigated in \cite{YC16a} -- we do not have a good regularization method for FFM, and hence need to rely on early-stopping to prevent over-fitting.
    We visualize this property in Figure \ref{fig:Overfitting}.
    To obtain the best test accuracy, the number of epochs must be carefully selected -- with insufficient epochs, the model can be under-fitting; on the other hand, with too many epochs, the model can be over-fitting.
    To determine the best number of epochs, we usually use a validation set to monitor the model performance at each epoch.
    Once the validation loss goes up, we stop the training process.
    We define three phases to indicate the ``maturity'' of the model.
    \begin{itemize}
        \item Pre-mature: the model is trained with too few epochs
        \item Mature: the model is trained with enough epochs
        \item Post-mature: the model is trained with too many epochs
    \end{itemize}

    \par
    The use of early stopping, however, makes warm-start difficult to be applied.
    If we seed a mature model to the next step and keep training, then the new model can be post-mature.
    This problem can be demonstrated in the following experiment.
    We again use Criteo's CTR Prediction Challenge dataset for reproducibility.
    We split the data set into 90 blocks, and at each step, 44 blocks are used for training, 1 block for validation, and 1 block for test.
    Therefore, the entire experiment starts from the 46th block (as test set), moves one block forward at each step, and ends at the 90th block (as test set).
    The validation set is used to determine the number of epochs.
    We first compare a \textit{baseline} setting, which did not use any warm-start approach, with a \textit{naive} warm-start described in Algorithm \ref{alg:NaiveSeeding}, which simply seeds the model obtained in the end of each step into the next step.
    
\begin{figure}[t]
      \centering
      \includegraphics[width=.5\textwidth]{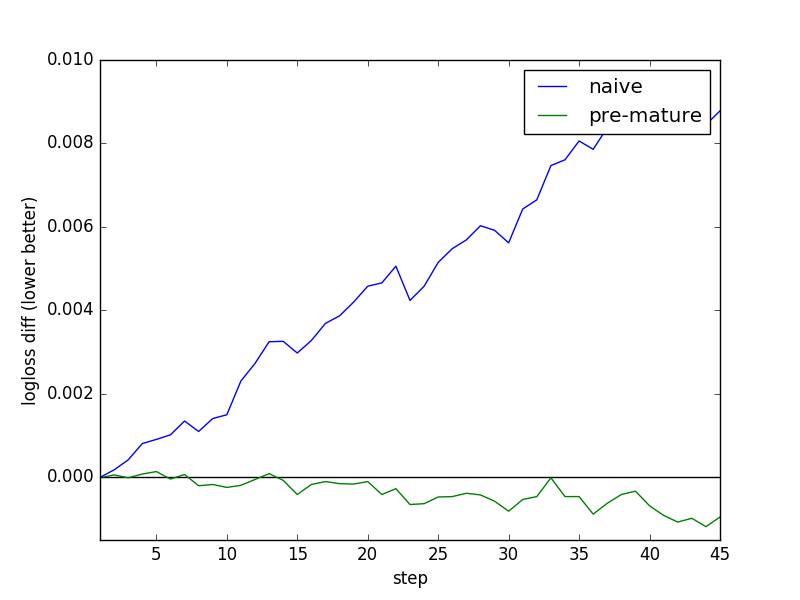}
      \caption{The test log loss of using FFM with different seeding approaches. The y-axis is the difference of log loss compared with the baseline (FFM without warm-start).}
      \label{fig:SeedingLogloss}
    \end{figure}

	The experiment result shown in Figure \ref{fig:SeedingLogloss} indicates that the post-mature problem indeed happens seriously -- the test accuracy is getting worse and worse when the experiments move forward.
    Again note that the goal of a warm-start technique is to reduce training time while keep the same predictability of the model.
    Clearly, by using a naive warm-start for FFM, this goal is not achieved.

    \par
    In this paper, we propose a new warm-start approach named \textit{pre-mature} warm-start.
    The idea is that instead of seeding a mature model to the next step, a pre-mature model is used as the seed.
    At each step, since the new model is initialized with a pre-mature model, it may be able to learn from the new data without over-fitting to the old data.
    For example, if the mature model comes at the 6th epoch, then this model will be used for prediction, but the model obtained at the 5th epoch will be seeded to the next step.
    The algorithm of pre-mature warm-start is described in Algorithm \ref{alg:pushback}.
    Here, ${\bf w}_{t-1}$ is used for prediction and ${\bf w}_{t-2}$ is seeded.

	\paragraph{Offline results}
	
\begin{figure}[t]
        \centering
      \includegraphics[width=.5\textwidth]{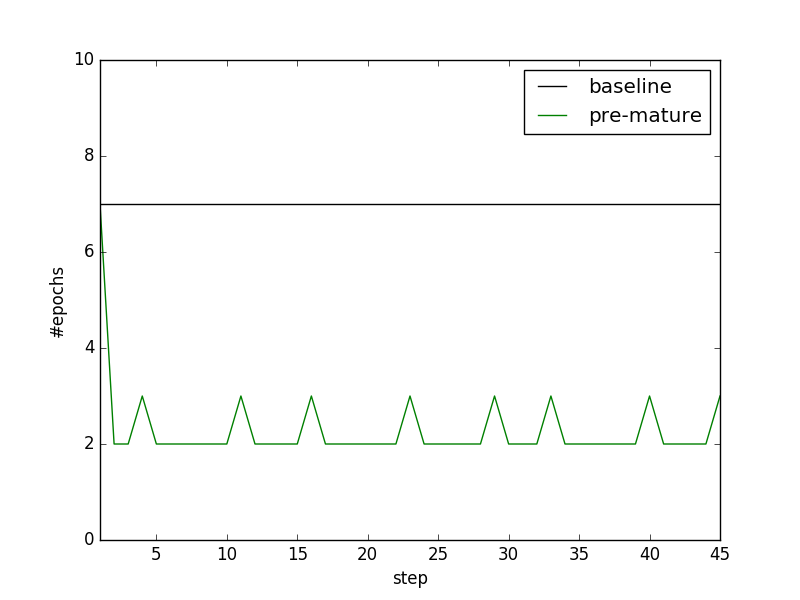}
        \caption{Number of epochs used in eash step. Both settings use 44 blocks of training data.}
      \label{fig:SeedingPushbackEpochs}
    \end{figure}

    The experiment results in Figure \ref{fig:SeedingLogloss} and \ref{fig:SeedingPushbackEpochs} show that with pre-mature warm-start, the test performance is not worse than the baseline any more, and the number of epochs required is significantly reduced.
    \begin{algorithm}[t]
        \begin{algorithmic}
        \Require an initial model ${\bf w}_{-1}$
          \State ${\bf w} \gets {\bf w}_{0} \gets {\bf w}_{-1}$
          \State calculate the validation loss $L_0$
          \For{$t \in \{1, \dots, T\}$}
            \State update ${\bf w}$
            \State ${\bf w}_t \gets {\bf w}$
            \State calculate the validation loss $L_t$
            \If{$L_t > L_{t-1}$}
              \State \Return (${\bf w}_{t-1}$, ${\bf w}_{t-2}$)
            \EndIf
          \EndFor
        \end{algorithmic}
        \caption{Our proposed ``pre-mature'' warm-start}
        \label{alg:pushback}
    \end{algorithm}
    \begin{figure}[t]
        \centering
      \includegraphics[width=.5\textwidth]{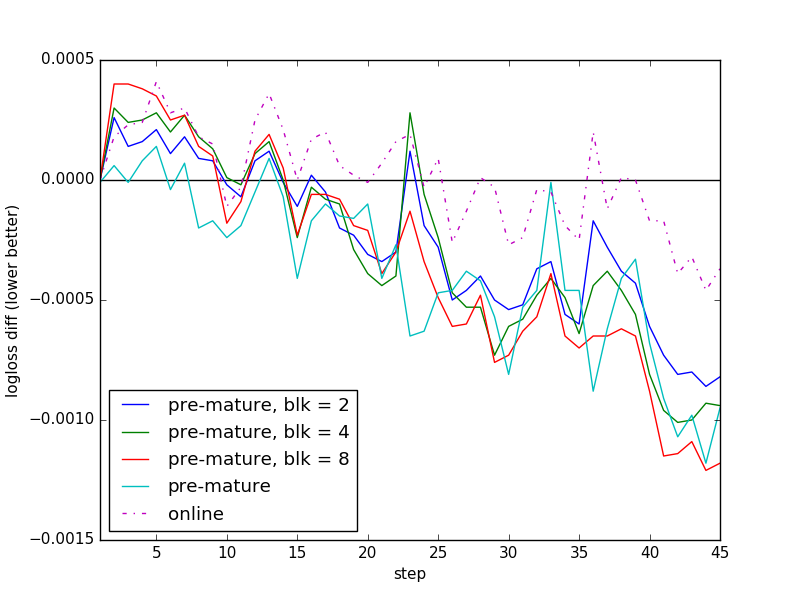}
        \caption{The log loss difference between baseline and different warm-start approaches and training sizes. The baseline (without warm-start) and pre-mature use 44 blocks as training data at each step. Note that we change the training size from the second step. For the first step, all settings use 44 previous blocks as training set, so the log loss are the same.}
      \label{fig:SeedingPrematureSize}
    \end{figure}

    \par
    It is noteworthy that the log loss of FFM with warm-start is getting lower as the experiment moves forward.
    This suggests that FFM may have some ability to remember the information learnt in the past.
    Inspired by this observation, we tried reducing the size of training set.
    Figure \ref{fig:SeedingPrematureSize} shows the comparison among different training sizes with pre-mature warm-start.
    We see that after sufficient number of steps, pre-mature with only 4 blocks of training set is still better than the baseline using 44 blocks.
    By using smaller training set, the training becomes much faster.
    The comparison of training time is shown in Table \ref{tab:ModelTrainingTime}.
    If we use 4 blocks for training, then it is 20 times faster than the baseline.
    \begin{table}[t]
        \centering
        \begin{tabular}{l|rrr}
                & \#epochs & time / epoch & total time \\
            \hline
            baseline           & 315 & 236s & 20.6hr \\
            pre-mature         & 103 & 236s &  6.8hr \\
            pre-mature (8  blks)  & 115 &  47s &  1.5hr \\
            pre-mature (4  blks)  & 128 &  26s &  0.9hr \\
            pre-mature (2  blks)  & 136 &  13s &  0.5hr \\
        \end{tabular}
        \caption{Total number of epochs, average time per epoch, and total training time of the entire experiment. Both baseline and pre-mature use 44 blocks as training data.}
        \label{tab:ModelTrainingTime}
    \end{table}

    \par
    An extreme case is to reduce the size of training set to only one block.
    In this case, because there is no overlap between two consecutive steps, we do not have to use pre-mature warm-start.
    (The purpose of pre-mature warm-start is to prevent over-fitting old data.) We illustrate this setting in the following figure, and refer to it as \textit{online}.
    \begin{center}
        \begin{tikzpicture}
            \draw [fill=gray] (2, 0) rectangle (3.0, -0.5);
            \node at (2.5, -0.25) {tr \#1};
            \draw (3.0, 0) rectangle (4.0, -0.5);
            \node at (3.5, -0.25) {va \#1};
            \draw             (4.0, 0) rectangle (5, -0.5);
            \node at (4.5, -0.25) {te \#1};

            \draw [fill=gray] (3, -0.5) rectangle (4.0, -1.0);
            \node at (3.5, -0.75) {tr \#2};
            \draw (4.0, -0.5) rectangle (5.0, -1.0);
            \node at (4.5, -0.75) {va \#2};
            \draw             (5.0, -0.5) rectangle (6, -1.0);
            \node at (5.5, -0.75) {te \#2};

            \node at (4.0, -1.25) {$\vdots$};
        \end{tikzpicture}
    \end{center}
    Figure \ref{fig:SeedingPrematureSize} shows FFM still can memorize the information under this setting, as it still out-performs the baseline.
    However, we do not use this setting because of two reasons.
    First, our proposed approach can achieve better log loss.
    Second, conceptually, if we only use a very small portion of data for training, the model can be very sensitive to the quality of this small set.
    Indeed, for example, at the 36th epoch in Figure \ref{fig:SeedingPrematureSize}, we see that online is worse than baseline while our proposed approach still out-performs baseline.

    \paragraph{Discussion}
    We have proposed two different ways to reduce training time.
    Distributed learning reduces the training time by adding more machines, but at the same time also {\em increases} the amount of computation.
    (In our previous experiments, when 32 machines are used, we needed roughly 3 times more epochs.)
    On the other hand, warm-start reduces the training time by initializing a model wisely and require less training epochs, which means the amount of computation is {\em decreased}.
    In a sense, warm-start seems to be a better approach than distributed learning.
    However, we cannot completely replace distributed learning with warm-start, because sometimes a cold-start is required, which means we need to train an entirely new model.
    In practice, this can happen when the code is updated or the system encounters unexpected error.
    In the cold-start scenario, we still need to rely on distributed learning to make sure we can learn the model on time.

\section{Conclusion}
\label{sec:conclusion}
    In this paper, we showed that Field-aware Factorization Machines can be successfully deployed in large scale advertising system, and that it significantly improves business metrics, in particular for small advertisers. One of the strengths of FFM is indeed their ability to generalize better than logistic regression through their use of a latent representation.
    
    Further, we proposed two ways to make training FFM faster: distributed learning and warm-start.
    The code for the experiments in Section \ref{sec:distributed} and \ref{sec:seeding} is available online.\footnote{\url{https://www.csie.ntu.edu.tw/~r01922136/ffmpaper2/exp/ffmpaper2-exp.tar}}
As future works, we plan to try our warm start method on our other non-convex problems that are difficult to regularize, such as a deep neural network.

\bibliography{sdp}
\bibliographystyle{abbrv}

\end{document}